# Machine Translation and Post-Editing: Comparative Evaluation of Different MT Systems and Post-Editor Groups in Specialised Translation

Joachim MINDER, Alexandra MESTIVIER, Natalie KÜBLER
Université Paris Cité, ALTAE, F-75013 Paris, France

This article aims to evaluate the quality of machine translation (MT) and post-editing (PE) in the context of specialised translation from English into French. Three MT systems (DeepL, eTranslation and Systran) were compared, and two groups of post-editors – linguists/translators and NLP experts – were asked to perform post-editing. Translation assessment is based on error annotation using an error typology adapted to MT and PE evaluation. The results reveal significant differences between the three MT systems and the two groups of post-editors, particularly in terms of terminological accuracy and fluency. This study highlights the importance of domain knowledge in specialised translation, as well as the limitations and variable performance of MT systems in language for specific purposes (LSP).

## 1. Introduction

Machine translation (MT) has become an increasingly widespread tool, used in a wide range of fields, including specialised translation. Advances in neural MT (NMT) systems have led to improvements in translation quality, making MT a standard tool in both professional and scientific contexts. However, the quality of the translations produced remains variable, often requiring human intervention through post-editing (PE). This activity, which involves revising and improving a machine-translated text, plays a key role in ensuring that the translated content meets specific requirements, particularly in specialised translation.

The rapid development of MT and PE raises a number of crucial questions: to what extent are existing MT systems adapted to specialised translation? To what extent do the profiles and skills of post-editors influence the final quality of the text? These questions are of particular relevance in the context of open science, where the accessibility of scientific publications depends to a large extent on high-quality translations. Research today is largely dominated by English (Swales, 1997; Gordin, 2015; Segue, 2023), which constitutes an obstacle for many non-English-speaking researchers and users. MT and MTPE offer a solution for improving accessibility to this knowledge by facilitating access to scientific articles and papers for an international readership. Scientific MTPE could be carried out either by domain experts or specialised translators, with the goal of making scientific research written in English accessible in other languages, or vice versa. However, for domain experts who lack translation expertise, the quality of their post-editing remains a critical issue. Conversely, for professional translators, the key challenge lies in their level of specialised knowledge. It is likely that these two groups face distinct challenges when post-editing MT output. Our experiment was conducted as part of the MaTOS (Machine Translation for Open Science) project (Bénard et al., 2023), which aims to develop MT methods adapted to scientific documents, to process term variation in MT, and to develop metrics for assessing the quality of the translations produced.[1]

MT and PE quality evaluation relies on different approaches. Automatic metrics, such as BLEU (Papineni et al., 2001), provide rapid and objective assessments by comparing a translation with a reference (meaning that translations produced by an MT system are compared "against human-translated references produced for the same source" (Albrecht & Hwa, 2008)), but these scores do not provide a qualitative and interpretable evaluation. Human evaluation, on the other hand, is often carried out using error annotation based on a typology, and offers a more detailed analysis of the strengths and weaknesses of translations, although it is more costly, cognitively demanding and time-consuming.

In this study, we evaluated the quality of machine-translated and post-edited texts in specialised translation, by comparing three MT systems (DeepL, eTranslation and Systran) and two groups of post-editors (linguists & translators, and experts in natural language processing). Our analysis was carried out with an annotation based on a typology of errors adapted to the evaluation of MT and PE, complemented by a parallel comparison with automatic metrics. The goal is to identify the strengths and limitations of each evaluation method, to observe how the post-editors' skills and profiles influence the final quality of the translations, and to determine whether the three MT systems provide translations of acceptable quality in Languages for Specific Purposes (LSP).

We first present an overview of the existing methods for evaluating translation, before describing our experimental protocol, the data used and our methods. We then present our results and analyse the performance of the different systems and groups of post-editors in specialised translation.

## 2. Review of Existing Evaluation Methods

Translation evaluation is a complex matter for which different methods exist, often serving different purposes. In general, the method used to evaluate translations depends on the nature of the translation, i.e. whether it is a human or machine translation. It is therefore important to distinguish between the two main types of translation: human translation (HT) and MT. Machine translation has made considerable progress, starting with rule-based machine translation in the early 1970s, followed by example-based machine translation and then statistical machine translation in the 1990s (Wang et al., 2022). Since the 2010s, a new MT paradigm has emerged: neural machine translation (NMT), a technology based on neural networks (Forcada, 2017; Vaswani et al., 2017). As NMT does not systematically produce translations of sufficient or acceptable quality, a new task has emerged at the intersection of HT and MT: post-editing (PE). This involves the modification, correction and revision by a human expert of a translation generated by an MT system (cf. e.g. Vaswani et al., 2017; Veale & Way, 1997) in order to obtain a translation of acceptable quality.

[1] Further information about the MaTOS project can be found here: https://anr-matos.github.io/en/index.html.

Automatic evaluation methods, referred to as metrics, are often used to evaluate MT, and sometimes also PE. Metrics are mainly used to evaluate MT systems in system development phase. Metrics compare machine translation with a reference translation based on strings of characters or words. The first metrics developed are referred to as superficial metrics, i.e. they focus solely on the superficial level of the sentence, without considering the full context (Nakhlé, 2023). Among these metrics is BLEU, which compares the n-grams (subsequences of *n* words) of a sentence in the translation with the n-grams of words in the sentence of the reference translation (Papineni et al., 2001). Another type of metric is learned metrics (also known as neural metrics), which are based on machine learning techniques. The most widely used is COMET (Rei et al., 2020), which is based on a multilingual model and which, unlike BLEU, also takes the source text into account. The advantage of metrics is that they are often quick to calculate and inexpensive. However, their scores are not appropriate for evaluating translations qualitatively, as they only calculate the similarity between a translation and a reference translation. Furthermore, most of the commonly used metrics do not take the source text into account.

In order to provide a more interpretable evaluation of a translation, it is often necessary to rely on human evaluation, mainly through annotation (Nakhlé, 2023). Error annotation consists of the labelling, by human experts, of the errors identified in a translation on the basis of an error typology. This method makes it possible to accurately identify the strengths and weaknesses of a translation, whatever its origin (HT, MT, or PE). The annotation is based on an error typology, i.e. a more or less granular framework that contains the different error types, classified into different categories. The most widely used typology in the world of MT assessment is the Multidimensional Quality Metrics (MQM) typology (Lommel et al., 2014).[2] This typology, which is intended to be universal, includes (to date) seven error categories: terminology, accuracy, linguistic conventions, style, locale conventions, audience appropriateness, design and markup; each of these categories being further divided into different error types for a total of more than 100 error types. A wide range of other typologies have been developed for specific purposes. For example, there are specialised typologies, i.e. typologies that focus on specific linguistic phenomena, such as Bénard (2019) and Kübler et al. (2022) for the translation of complex noun phrases. Other typologies have also been developed for translation training, such as MeLLANGE (Castagnoli et al., 2011; Kübler, 2008), initially developed for annotating HT and then later adapted for MT and PE annotation (Kübler et al., 2021). Although no typology is generally accepted in the community, the various existing typologies share the same categories or even sub-categories of errors, which are, however, organised or named in different ways. Some typologies are more granular than others, granularity being a relevant issue, since it is important to maintain a balance between a typology that is too granular – which can be difficult to manipulate – and one that is too condensed – which can be imprecise. Human evaluations are more expensive than metrics, are more time-consuming and require more demanding cognitive efforts. In addition, human resources cannot be re-used (Papineni et al., 2001), i.e. human evaluations cannot be repeated in a systematic and automatic way.

---

[2] More information about MQM can be found here: https://themqm.org/error-types-2/typology/.

Error annotation can also be used to evaluate post-editing, provided that the typology considered is adapted to PE evaluation. Lefer et al. (2022) have proposed the MTPEAS (Machine Translation Post-Editing Annotation System) typology, which consists of a decision tree that defines the effect of the modifications made by the post-editor on the quality of the translation (incorrect modification, superfluous modification, good correction, etc.). This typology is therefore not based on linguistic categories of error, but on categories of modification type. The previously mentioned MeLLANGE typology (Castagnoli et al., 2011; Kübler, 2008; Kübler et al., 2022) combines this decision tree approach for PE evaluation with annotation of linguistic errors, which means that MeLLANGE is appropriate for the evaluation of human translation, MT and PE. However, post-editing can also be evaluated automatically. For example, the HTER measure (Snover et al., 2006) is used to calculate the number of edits made in post-editing to fix the corresponding MT output.

This review shows that translation evaluation methods differ according to the evaluation objective and the translation type (HT, MT, PE). In the next section, we present our experimental protocol and the methods for evaluating machine-translated texts and post-edits.

## 3. Experiment

This study aims to evaluate the quality of machine-translated texts and post-edits in the field of natural language processing (NLP). The experiment involves the post-editing of MT output generated by three different MT systems: DeepL, eTranslation and Systran. These systems, which are widely used in various professional and academic contexts, were selected in order to investigate their respective performance in specialised translation. One of the main goals of this experiment is to analyse the quality of the translations produced by these MT systems when used for technical and scientific translation in the field of NLP. By comparing their performance, we can identify the strengths and weaknesses of each system. In addition, our experiment focuses on the differences between two groups of post-editors: on the one hand, NLP experts, who are expected to have a solid knowledge of the content and concepts specific to NLP, and on the other, linguists and translators, who have a solid knowledge of cross-linguistic transfer and the linguistic and stylistic requirements of a translation.[3] The analysis focuses in particular on each group's approach to post-editing, the types of errors that post-editors most frequently make and the nature of the modifications made to machine-translated texts. By combining the evaluation of MT systems and the study of the differences between the two profiles of post-editors, this experiment sheds light on the quality of machine translations in specialised fields and on the impact of post-editing in improving these translations. It also enables us to examine the extent to which domain or linguistic expertise influences the way in which errors in an MT text are perceived and corrected.

[3] We believe that there is a significant distinction to be made between linguists and translators. However, as not all subjects are translators (there are also professors in the field of translation studies and linguistics), nor are all subjects strictly speaking linguists (e.g. translation students), we have decided, for the sake of clarity in this article, to refer to this group of post-editors as ‘Translators’, although we are aware that not all of them are translators.

## 4. Methods

### *4.1. Corpus of source texts*

The corpus of source texts contains abstracts of scientific publications in English in the field of NLP, together with the corresponding titles, retrieved from the HAL open archive.[4] They mainly consist of conference papers, scientific journal articles, books, preprints, reports or book chapters. The source texts were selected through qualitative filtering. Initially, only publications related to the general field of computer science were selected. Next, a second filter was applied to retain only publications dealing with NLP on the basis of keywords. Finally, publications with an abstract translated into French and publications that were not open access were excluded.

### *4.2. Corpus of MT outputs*

These English titles and abstracts were subsequently translated into French using three commercial MT systems: DeepL Pro (version 7.5), Systran Translation Pro and eTranslation (version 12.3), all of which are NMT systems.

### *4.3. Corpus of post-edits*

Two distinct groups of subjects were involved in the (French) post-editing of these machine-translated texts: (a) experts in the field of NLP, who were encouraged to post-edit their own publications (referred to as 'NLP experts' in the following), and (b) a more heterogeneous group of translators, linguists, students and professors of the Master's programme in specialised translation at Université Paris Cité (referred to as 'Translators' in the following). The post-editors included four Translators (three of them native French speakers) and 16 NLP experts (13 of them native French speakers) with experience in NLP ranging from 3 years to more than 10 years.

Each abstract had a maximum of three post-edits, corresponding to one post-edit for each MT system, but the data related to the MT system was not made available to the post-editors during the post-editing process. All post-editors were given the same instructions (in French, translated here into English):

> Edit the text (title and abstract) to make it clear, understandable and acceptable, as it should be done for a publication in a French-language journal (e.g. the journal 'TAL'). To do so, please refrain from using machine translation tools. If possible, please complete this task without interruption, so that the time recorded corresponds to the actual post-editing time.

As indicated, the time needed by the post-editors was recorded. However, this data may have been biased if the post-editors interrupted themselves during their task, although this was explicitly discouraged. After each post-edit, the post-editors were asked questions. For the NLP experts, the question was: ‘How important would you consider the translation issues identified? They could answer ‘no problem identified’, ‘minor’, ‘moderate’ and ‘major’. The Translators were asked more specific questions; they were asked to rate the severity of errors in the following categories: faithfulness to source text, grammar, terminology, spelling and punctuation, style as well as textual coherence. They were provided with the same four levels of severity.

[4] The HAL open archive can be found here: https://hal.science.

**Table 1.** Corpus data: total number of texts contained in the corpus and total number of texts (source, MT and PE) + words (WC, word count) annotated for the purposes of human evaluation

<table>
<tr><td></td><td colspan="2">NLP experts PE</td><td colspan="3">Translators PE</td><td colspan="2">Total</td></tr>
<tr><td>Total of abstracts post-edited</td><td colspan="2">95</td><td colspan="3">242</td><td colspan="2">337</td></tr>
<tr><td colspan="8"></td></tr>
<tr><td rowspan="2">Abstracts post-edited by both groups (i.e. annotated manually)</td><td>#</td><td>MT WC total // avg.</td><td>Source WC total // avg.</td><td colspan="3">NLP experts PE WC total // avg.</td><td>Translators PE WC total // avg.</td></tr>
<tr><td>24</td><td>4.205 175,21/MT</td><td>3.609 // 150,38/abstract</td><td colspan="3">4.696 // 195,67/PE</td><td>4.259 // 177,46/PE</td></tr>
</table>

*4.4. Parallel corpus annotation*

The aligned parallel corpus obtained consists of the English source texts, the MT output and the post-edits by the two groups. Each MT output is associated with the system which performed the translation; for each post-edit, the profile of the post-editor (NLP expert or Translator) is also indicated. In our corpus, three types of post-edits are available: post-edits carried out by NLP experts only, post-edits carried out by Translators only, and texts for which we have both NLP expert and Translator post-edits. For the purposes of this study, we have annotated the post-edits carried out by both audiences only, so as to allow us to compare similar data. On the other hand, for the automatic evaluation (metrics), all the texts were evaluated.

To annotate the corpus, we used brat rapid annotation tool, which allows annotators to add predefined labels to segments of the corpus (see

Figure ***1***).[5]

Figure 1 shows that the corpus is aligned at the sentence level, with one line for the source text, one line for the MT output and one line for the PE.

[5] Find out more about the annotation tool here: https://brat.nlplab.org/index.html.

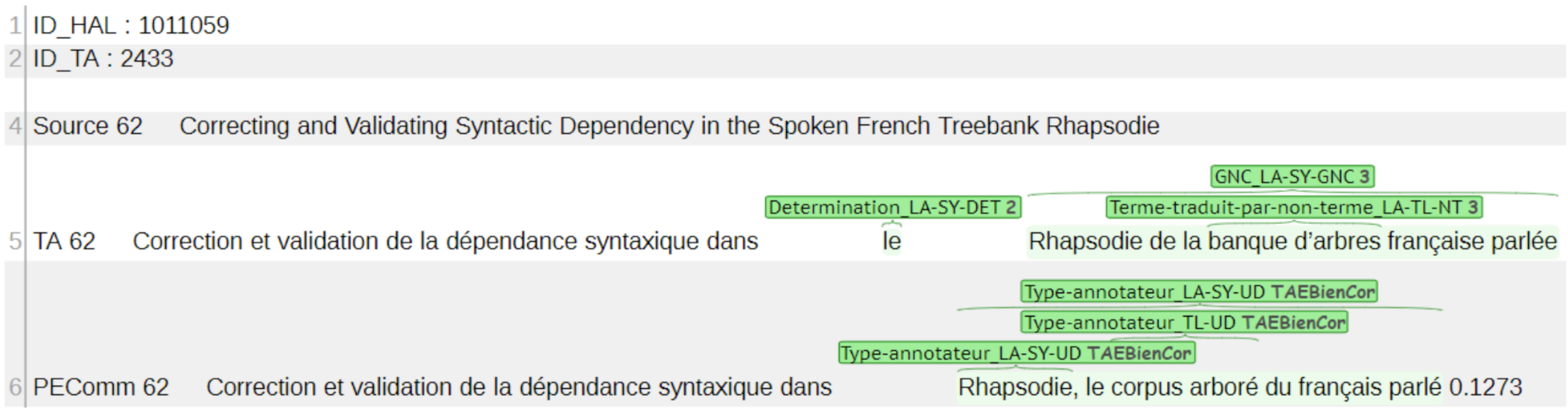

**Figure 1.** A view of the brat program used to annotate the corpus

*4.5. Error typology used*

Our corpus was annotated on the basis of an error typology adapted for the purposes of this experiment. Our starting point was the MeLLANGE typology (Castagnoli et al., 2011; Kübler, 2008) and its enhanced version (Kübler et al., 2021), to which we added some error types from the MQM typology (Lommel et al., 2014), since some categories were missing from the MeLLANGE typology but were included in the MQM typology. Our full typology is presented in Appendix 1 (
Figure 7).

Our error typology is divided into three main error categories: content transfer errors, language errors and tool-related errors. The 'content transfer' category covers errors known as translation errors, i.e. errors that alter the meaning and content of the source text or affect the transfer and comprehension of the message. This broad category includes errors such as omissions, additions, content distortions, indecisions, source text intrusions and target language intrusions. The 'language' category includes linguistic errors. These include errors of syntax, inflection and agreement, typography, register, style, reference, textual conventions, as well as a wide range of terminological errors (LSP) and lexical errors (Language for General Purposes). Finally, the 'tools' category, which was introduced for the purposes of this project, covers errors related to tools or the command of tools. These include MT hallucinations, non-compliance with the corpus or glossary provided - where applicable - and duplication errors. All these errors are used to annotate both MT and PE.

In addition to this error typology, we use a number of attributes to annotate PE only (Kübler et al., 2021). These attributes are used to describe the relationship between the MT output and post-editing. We use five different attributes which can be represented as follows:

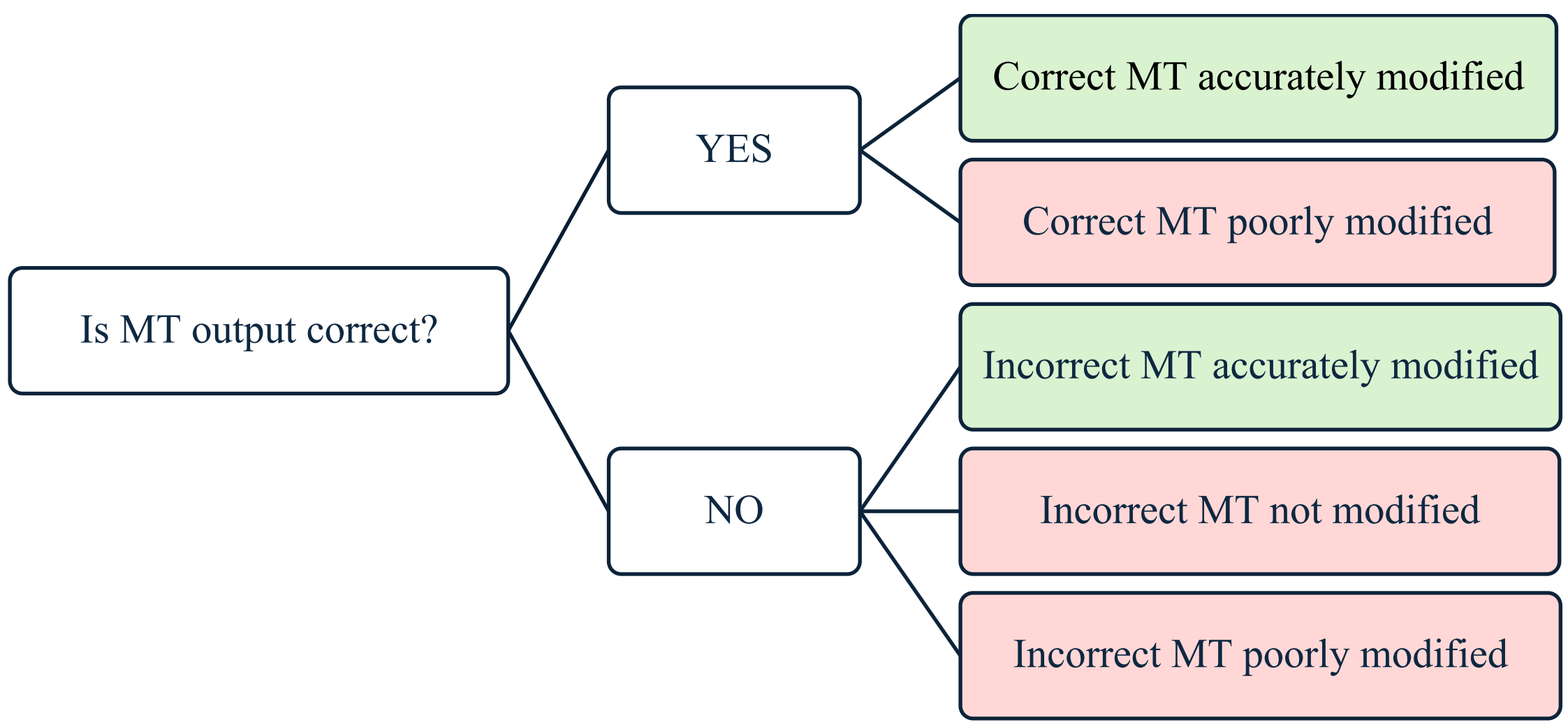

**Figure 2.** Decision process for annotating post-editing attributes; attributes on a green background are not counted as errors, while attributes on a red background are counted as errors

Finally, for all errors (in MT and PE), a severity score is assigned to each error. This is a widely used annotation method, particularly in the MQM framework, which provides an example of a scorecard, i.e. an evaluation sheet for a translation, where four levels of severity are employed: neutral error (score 0), minor error (score 1), major error (score 5), critical error (score 25).[6] We have chosen to apply these severity levels. A neutral error means that a better translation could be provided, but that the translation is not truly erroneous. A minor error has a (very) slight impact on the target text, but does not affect the readability or comprehension of the content. A major error has a more significant impact on the target text, i.e. it affects its comprehension, readability or relevance. Finally, an error is considered critical if it renders the content completely false or unexploitable, i.e. total rewriting is required.

*4.6. Scorecards*

Following the corpus annotation, the data was exported and processed. For each source text, an Excel spreadsheet was created with the annotated data of the MT output, the post-edited data of NLP experts and the post-edited data of Translators, in order to create scorecards. Scorecards are designed to evaluate the overall quality of a MT text or a PE (see Table 2 for an example of a scorecard).

[6] You can get more information about MQM scorecards here: https://themqm.org/error-types-2/the-mqm-scoring-models/.

**Table 2.** Example of a scorecard created with annotated data from a machine-translated text by eTranslation

| MT (eTranslation) | Error total | 0 | 1 | 5 | 25 | Severity total |
|---|---|---|---|---|---|---|
| Content-transfert | 0 | 0 | 0 | 0 | 0 | 0 |
| Omission_TR-OM | 1 | 0 | 1 | 0 | 0 | 1 |
| Source-language-intrusion | 0 | 0 | 0 | 0 | 0 | 0 |
| Untranslated-translatable_TR-SI-UT | 1 | 0 | 1 | 0 | 0 | 1 |
| Too-literal_TR-SI-TL | 1 | 0 | 1 | 0 | 0 | 1 |
| Language | 0 | 0 | 0 | 0 | 0 | 0 |
| Syntax_LA-SY | 0 | 0 | 0 | 0 | 0 | 0 |
| Determination_LA-SY-DET | 1 | 0 | 1 | 0 | 0 | 1 |
| Wrong-preposition_LA-SY-PR | 1 | 0 | 0 | 1 | 0 | 5 |
| Complex-NP_LA-SY-GNC | 4 | 0 | 1 | 3 | 0 | 16 |
| Inflection-agreement | 0 | 0 | 0 | 0 | 0 | 0 |
| Number_LA-IA-NU | 1 | 0 | 1 | 0 | 0 | 1 |
| Typography | 0 | 0 | 0 | 0 | 0 | 0 |
| Spelling_LA-HY-SP | 1 | 0 | 1 | 0 | 0 | 1 |
| Style | 0 | 0 | 0 | 0 | 0 | 0 |
| Awkward_LA-ST-AW | 1 | 0 | 1 | 0 | 0 | 1 |
| Textual-conventions | 0 | 0 | 0 | 0 | 0 | 0 |
| Cohesion_LA-TC-CN | 1 | 0 | 1 | 0 | 0 | 1 |
| Terminology-and-lexis | 0 | 0 | 0 | 0 | 0 | 0 |
| Incorrect-choice-terminology_LA-TL-INS | 4 | 0 | 3 | 1 | 0 | 8 |
| Term-translated-by-non-term_LA-TL-NT | 4 | 0 | 4 | 0 | 0 | 4 |
| Tools | 0 | 0 | 0 | 0 | 0 | 0 |
| | 21 | | | | | 41 |
| | | | | | | |
| Word count | 147 | | | | | |
| Penalty/100 words | 27,89115646 | | | | | |
| **Quality Score** | **72,1088435** | | | | | |

A quality score is computed on a total of 100 points. To compute this score, a severity total must be calculated first (see column on red background in Table 2), which is first calculated for each type of error with the different severity levels. The total severity score is then calculated for 100 words, using the word count of the source text as a reference, resulting in the penalty/100 words, calculated as follows:

$$\frac{severity\ total}{word\ count\ of\ source\ text}\ x\ 100$$

Finally, to compute the overall quality score, the penalty/100 words must be subtracted from 100 (score corresponding to a perfect translation).

## 5. Results

The corpus was evaluated in two different ways: through automatic metrics and through human annotation. For the automatic evaluation, the whole corpus was considered, while only the MT outputs post-edited by both post-editor groups were annotated.

### 5.1. Results of the automatic evaluation

The automatic evaluation is intended to examine the following four questions: (1) How much effort is required by the two different groups of post-editors to perform the PE of MT output? (2) Is it possible to measure the quality of these post-edits? (3) Do the different MT systems (DeepL, Systran, eTranslation) show different results in terms of quality? (4) What types of errors cause the main problems in this process?

#### 5.1.1. Measuring post-editing effort

The necessary post-editing effort was calculated using the HTER metric (Snover et al., 2006), which calculates the distance in terms of edits between post-editing and machine translation (number of edits needed to go from MT to PE). This metric was calculated for each abstract individually, and then averaged over each group of post-editors.

**Table 3.** Average HTER score per document for each post-editor group

| HTER (Human-targeted Translation Edit Rate) score | |
|---|---|
| Average for NLP experts | 18.2 |
| Average for Translators | 8.0 |

Table 3 shows that the HTER score is relatively low, suggesting that MT outputs are generally of high quality as they require little post-editing effort. Of the 337 documents, 13 were left completely unedited. However, there was a difference of more than 10 points between the HTER scores of the NLP experts and the Translators, indicating that more post-editing effort was required for the NLP experts. This difference could reflect the way in which each group perceives post-editing. Indeed, as members of the translation community (translation students and teachers, translators, etc.) are expected to be familiar with post-editing, they tend to follow established post-editing guidelines, i.e. only make necessary changes (cf. e.g. Taus, 2010). Another possibility is that they lack the specialised knowledge required to identify the errors.

#### 5.1.2. Measuring the quality of post-edits and MT outputs

To measure the quality of post-edits and MT outputs without reference translation, we used the COMET metric (Rei et al., 2020), a neural metric based on a multilingual model. The results show that there is little to no significant improvement between the quality of MT outputs and the quality of PEs, for both groups of post-editors (see Table 4).

**Table 4.** Average COMET scores for MT and PE for NLP experts and Translators

| | MT | PE | Increase |
|---|---|---|---|
| COMET score for NLP experts | 77.8 | 78.6 | + 0.8 |
| COMET score for Translators | 76.3 | 77.0 | + 0.7 |

Although the difference is only marginally significant, the quality of the post-edits by NLP experts is higher than the quality of the post-edits by Translators, almost proportionally to the quality of the MT outputs edited by the two groups.

*5.1.3.Comparing MT systems*

To compare the performance of the three MT systems, two metrics were used: HTER and the BLEU score, designed to measure the degree of similarity between MT and PE.

Table 5 shows that the MT system evaluated as the most performing is DeepL, followed by Systran and then eTranslation. The lower the BLEU score (the poorer the MT is perceived to be), the higher HTER (the greater the post-editing effort), as expected. On the other hand, all three MT systems can be considered as performing reasonably well for specialised translation in NLP, none of them achieving a dramatically low BLEU score.

**Table 5.** Comparison of BLEU scores and HTER for each MT system according to post-editor groups

| Group | DeepL | Systran | eTranslation |
|---|---|---|---|
| NLP experts | 0.817 / 13.8 | 0.731 / 19.7 | 0.683 / 24.4 |
| Translators | 0.904 / 7.1 | 0.873 / 8.5 | 0.859 / 10.6 |
| *Average* | *0.861 / 10.5* | *0.802 / 14.1* | *0.771 / 17.5* |

*5.1.4. Error analysis*

An analysis of the post-editors' answers to the questions asked after each post-editing session reveals that errors in grammar, style, punctuation, faithfulness to the source text and consistency were generally associated with low severity levels. On the other hand, terminological errors are often associated with higher levels of severity, which attests to the specialised nature of our corpus.

*5.2. Results of the human evaluation*

The scorecards created after annotating the corpus were used to provide statistics on a number of areas of analysis: the comparative overall quality of MT outputs and post-edits, error types, severity scores and post-editing attributes.

*5.2.1. MT and PE quality*

The automatic metrics showed very little improvement between the quality of MT and PE for the two groups of post-editors, as well as a fairly minor difference in performance between the three MT systems. The human evaluation reveals the same trend in terms of MT system ranking, but with more marked differences between systems.

Figure ***3*** shows that the system with the highest score is DeepL, followed by Systran, and then eTranslation, which matches the system ranking by metrics. However, the gap between the two highest scoring systems and eTranslation is statistically significant (+50 pts), suggesting that eTranslation's performance is particularly weak. These patterns are not entirely reflected in the quality of post-edits. Systran is the system whose post-edits are rated as the best (89.68 on average). Although DeepL provides the highest scoring MT outputs, its post-edits are slightly lower quality (84.53 on average) than Systran's post-edits. The quality of eTranslation's post-edits, on the other hand, varies considerably depending on the group of post-editors; for NLP experts, the low quality of eTranslation's MT does not seem to influence the quality of their post-edits, unlike Translators, whose post-edits of eTranslation have a fairly low score. For Translators, there seems to be a correlation between the performance of the MT system and the quality of the post-edits, i.e., the lower the quality of the MT system output, the poorer the post-editing results.

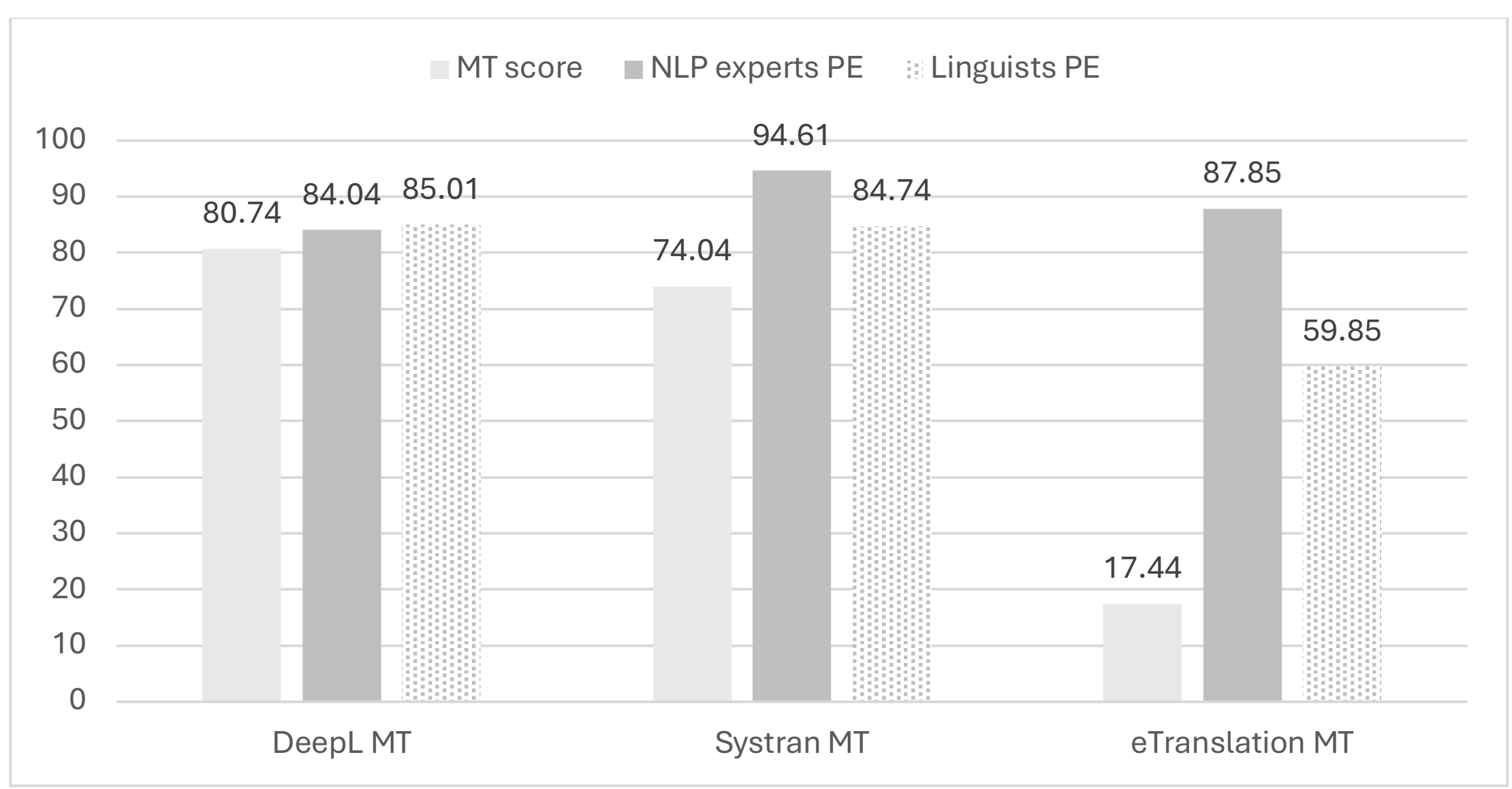

**Figure 3.** Distribution of average scores by MT system and post-editor group

On average, NLP experts produce post-edits that are judged to be better than the post-edits produced by Translators (see Table 6).

**Table 6.** Average scores for MT outputs and PE for both groups and degree of improvement between MT and PE

| MT average score | 61.67 |
|---|---|
| NLP experts PE average score | 87.22 (+ 25.55) |
| Translators PE average score | 77.91 (+ 16.24) |

The results shown in Table 6 partially confirm the findings regarding the metrics. However, the human evaluation reveals significant improvements between the quality of MT and PE, with a greater improvement for NLP experts.

*5.2.2. Statistics on post-editing attributes*

Table 7 shows that NLP experts have the highest proportion of accurately corrected errors (Incorrect MT accurately modified) and the highest proportion of the introduction of correct variants (Correct MT accurately modified). This echoes one of the observations made with the metrics, which showed that NLP experts were more likely to introduce modifications into their PE (higher HTER). Furthermore, Translators identified and corrected fewer errors than NLP experts (Incorrect MT not modified). While the difference between the two groups is not statistically significant, it can be seen that NLP experts introduce the most errors in their post-edits (Correct MT poorly modified), i.e. they are more likely to modify a correct MT and thus introduce an error in the PE.

**Table 7.** Distribution of PE attributes according to post-editor groups

| Attributes | NLP experts PE | | | Translators PE | |
|---|---|---|---|---|---|
| Incorrect MT accurately modified | 255 | 52.58% | | 194 | 44.80% |
| Correct MT accurately modified | 99 | 20.41% | | 49 | 11.32% |
| Incorrect MT poorly modified | 30 | 6.19% | | 33 | 7.62% |
| Incorrect MT not modified | 68 | 14.02% | | 131 | 30.25% |
| Correct MT poorly modified | 33 | 6.80% | | 26 | 6.00% |
| **Total errors** | **131** | | | **190** | |

In total, NLP experts made 131 errors, while translators made 190, hence the difference in quality between the two groups of post-editors.

*5.2.3. Statistics on severity levels*

Figure 4 shows the distribution of severity levels (0, 1, 5, 25) according to the MT system.

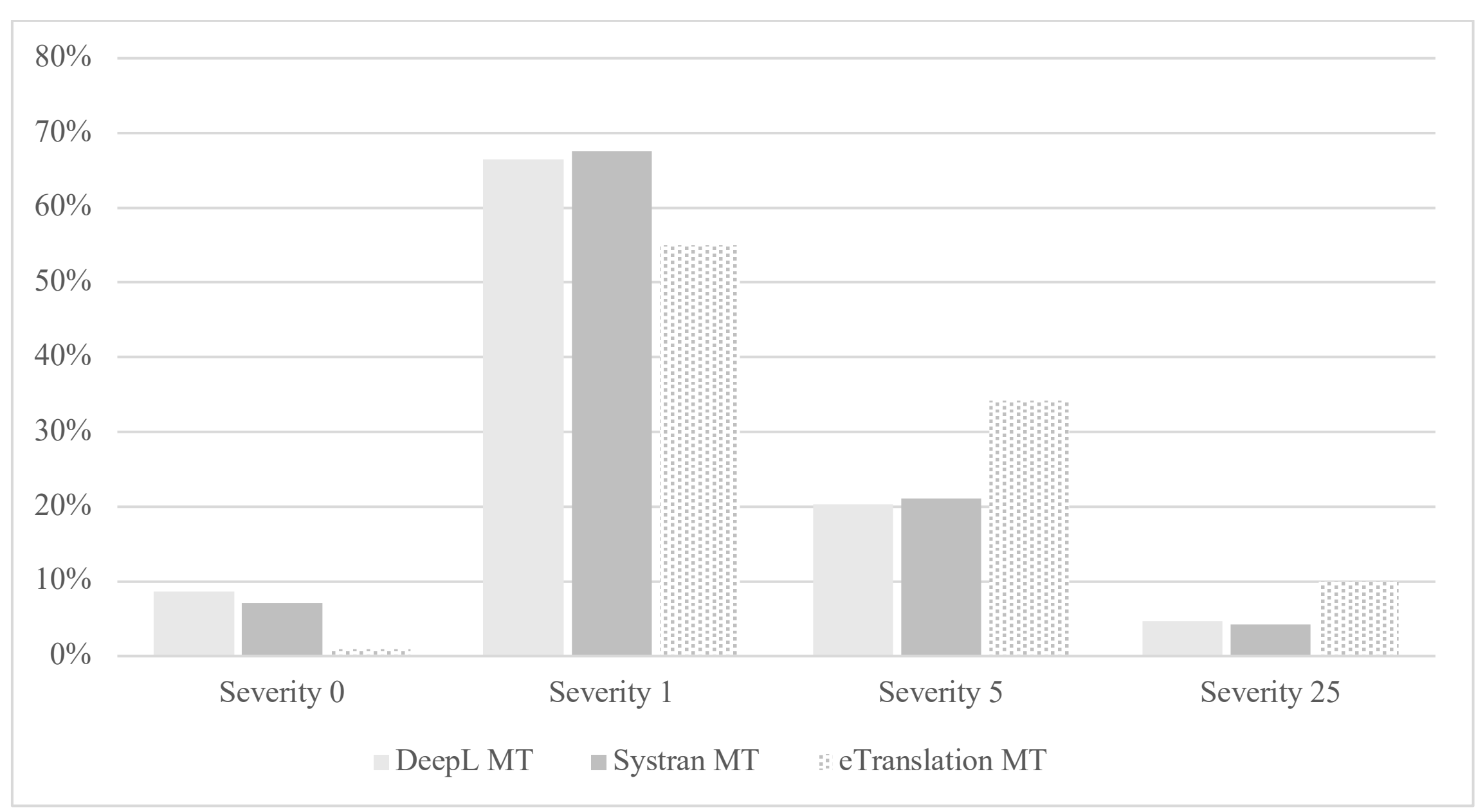

**Figure 4.** Distribution of severity levels according to MT systems

Figure 4 may partially illustrate why eTranslation is the MT system with the lowest quality scores. We can observe that eTranslation makes a larger proportion of major and critical errors (levels 5 and 25) than the other systems, and these weights account for a significant proportion of the quality scores. eTranslation also generates fewer neutral errors (score 0) and minor errors (score 1). On the other hand, DeepL and Systran have comparable profiles in terms of the distribution of severity scores.

Figure ***5*** shows the same distribution of error weights, but according to the group of post-editors.

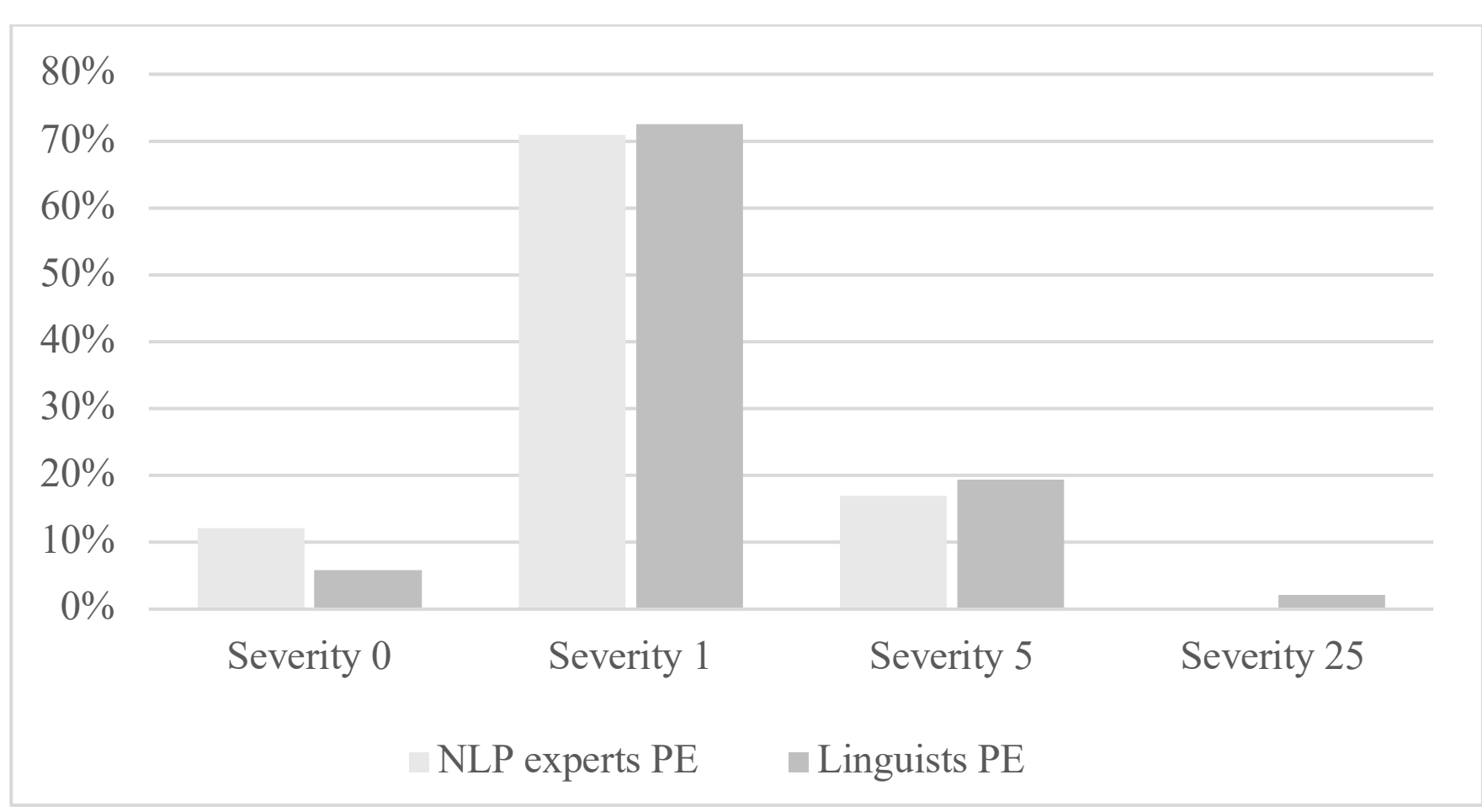

**Figure 5.** Distribution of severity levels according to post-editor groups

Figure 5 reveals that the two groups of post-editors have roughly similar profiles in terms of severity score distribution, with a few insignificant discrepancies. Firstly, it can be noted that there are fewer major and critical errors in the post-edits of the two groups than in the MT outputs (all systems considered), which may explain the improvement in the quality score between the MT and the PE. NLP experts make more neutral errors (score 0), and Translators make more 'genuine' errors (minor, major and critical), which explains the slightly lower quality of the Translators' post-edits. However, it should be noted that the statistics on the severity score distribution between NLP experts and Translators must be treated with caution, given that all the critical errors (score 25) are identified in the same post-edit of a single Translator, suggesting that the issue may come from a single post-editor and not from the Translator group as a whole.

*5.2.4. Error analysis*

Analysis of the automatic metrics (answers to the questions after post-editing) suggested a prevalence of terminology errors, which were often associated with higher levels of severity. It is therefore interesting to examine the proportion of terminology errors in relation to other types of error and the average number of terminology errors per text depending on the MT system and the post-editor group.

**Table 8.** Number of terminological errors per text and proportion of terminological errors in relation to other types of error, according to MT systems and post-editor groups

| | # of terminology errors/text | proportion of terminology errors compared to other errors |
|---|---|---|
| DeepL MT | 4.5/text | 46.09% |
| Systran MT | 6/text | 42.86% |
| eTranslation MT | 8.1/text | 35.85% |
| NLP experts PE | 1.6/text | 29.08% |
| Translators PE | 2.6/text | 32.18% |

Table 8 shows that among MT systems, DeepL produces the fewest terminological errors per text, while eTranslation produces the most per text. On the other hand, in terms of the proportion of terminology errors in relation to other error types, for DeepL and Systran, terminological errors represent almost half of all errors generated, while the figure is lower for eTranslation. This suggests that eTranslation makes more terminology errors than the other two systems, but that it also has a higher number of other error types, hence the lower proportion. As far as post-editors are concerned, NLP experts make few terminology errors (1.6/text), while Translators make slightly more (2.6/text), although this number is still lower than for MT systems, suggesting that most terminology errors are corrected by both groups.

We also analyse the 5 most frequent errors for each MT system. Table 9 provides an overview of the five most frequently identified errors for the three MT systems. Each error type represented in the table is defined in Appendix 2 (Table 11). Two types of error are shared by all three systems. These are the errors of wrong choice of terminology and of term translated by non-term, both of which are terminology errors, as would be expected given the statistics presented in Table 8, which is also in line with previous results with the same kind of error annotation in LSP translation. However, Table 9 also shows the error types that are specific to certain MT systems. For example, Systran is the only system to have style and register errors among its 5 most frequent errors; this may suggest that Systran produces translations which are less natural from a stylistic point of view in the target language than the other two systems. eTranslation has another specificity: it is the only system to have preposition errors among its 5 most frequent errors. For a more detailed analysis and ranking of the most frequent errors for each MT system, see Table 11 in Appendix.

**Table 9.** Overview of the five most frequent errors for the three MT systems (marked with •)

| Error types | DeepL MT | Systran MT | eTranslation MT |
|---|---|---|---|
| Awkward style | | • | |
| Complex noun phrase | • | | • |
| Determination | • | • | |
| Distorsion | • | | • |
| Inappropriate register for target text | | • | |
| Incorrect terminological choice | • | • | • |
| Term translated by non-term | • | • | • |
| Wrong preposition | | | • |

Table 10 presents the same frequent error statistics, but for post-editors. Two error types are common to both groups: distortion and incorrect terminological choices, in line with translation students in specialised human translation (Kübler et al., 2016a; 2016b). Regarding the specificities of each group, NLP experts are more likely to make awkward style errors, omissions and overly literal translations, whereas Translators tend to make more errors in translating complex noun phrases, determiners and to translate terms with non-terms. In principle, Translators appear to make more language errors (and terminology errors, as discussed above) than NLP experts, who in turn seem to make more content transfer errors. For a more detailed analysis and ranking of the most frequent errors for each post-editor group as well as definitions for each error type, see Table 11 in Appendix 2 for a more detailed analysis.

**Table 10.** Overview of the five most frequent errors for both groups of post-editors (marked with •)

| Error types | NLP experts PE | Translators PE |
|---|---|---|
| Awkward style | • | |
| Complex noun phrase | | • |
| Determination | | • |

| Distorsion | • | • |
|---|---|---|
| Incorrect terminological choice | • | • |
| Omission | • | |
| Term translated by non-term | | • |
| Too literal | • | |

One final statistic we present relates to the distribution of error categories across translation systems and post-editor groups.

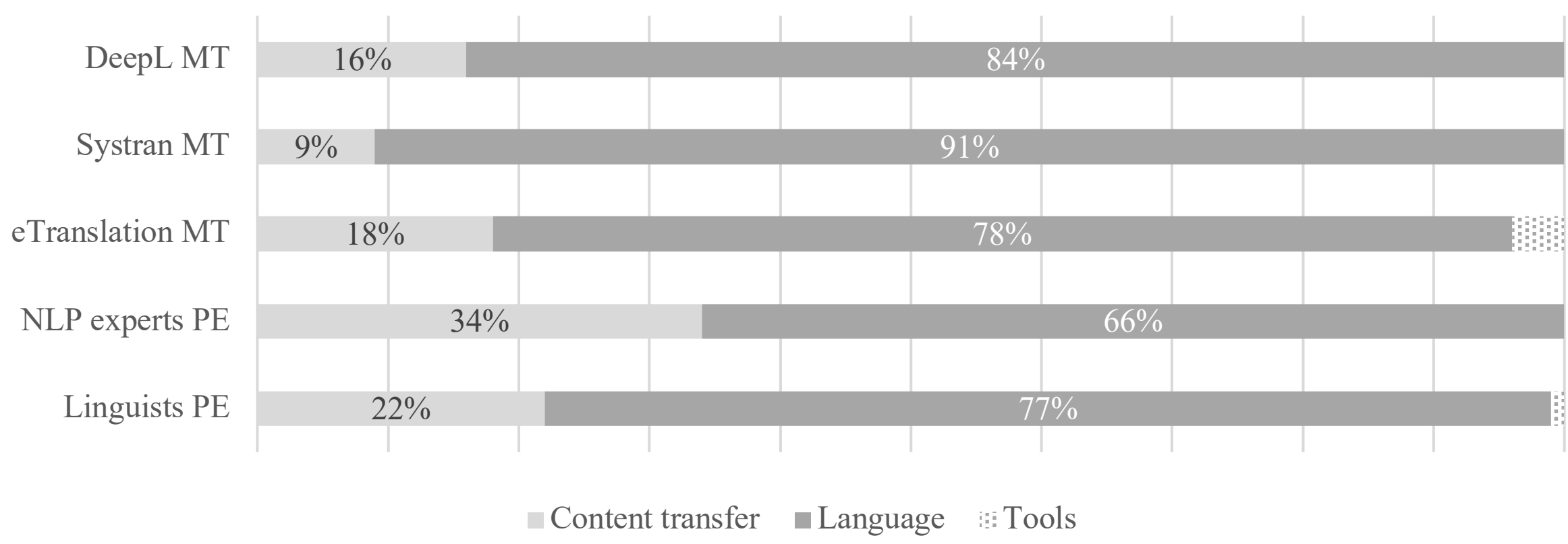

**Figure 6.** Distribution of error categories across MT systems and post-editor groups

Figure 6 reveals nuances in the error profiles of the different MT systems and post-editor groups. For MT systems, DeepL and eTranslation have fairly similar profiles in terms of content transfer and language error distribution. However, eTranslation is the only MT system to produce tool-related errors, in this case hallucination errors, a critical error by definition (see Table 11 in Appendix 2 for the definition). On the other hand, Systran follows a similar trend, but produces translations with a meaning that is closer to the source texts (fewer content transfer errors), but produces more language and terminology errors. Turning to the profiles of the two groups of post-editors, Translators seem to make more language (and terminology) errors than NLP experts, who produce post-edits that are more accurate linguistically and terminologically, but more imprecise in terms of content transfer. Translators are the only group to make tool-related errors (1%).

## 6. Discussion

The first goal of this experiment was to compare the scores of the automatic metrics (BLEU, HTER and COMET) with the statistics of the human evaluation. In terms of MT and PE quality, the metrics do not fully coincide with the annotation results. The two evaluation methods agree on the system ranking, with DeepL producing the best translations on our corpus, followed by Systran and then eTranslation. However, the automatic evaluation revealed relatively minor differences between the performances of the three MT systems, whereas the annotation showed a particularly significant difference in quality between Systran and eTranslation, the latter performing particularly poorly on our corpus of NLP texts. When it comes to post-edits, the two types of evaluation differ. Based on metrics, post-edits of DeepL MT were the highest rated, followed by Systran's and then eTranslation's. However, according to the human evaluation statistics, the post-edits of Systran MT were of the highest scoring, followed by those of DeepL and then those of eTranslation. The automatic metrics also show a very slight improvement between MT and PE for both groups of post-editors (less than 1 point on average). On the other hand, human evaluation shows much more significant improvements between MT and PE (+26 pts on average for NLP experts and +16 pts on average for Translators). However, the two types of evaluation do coincide in one respect: NLP experts post-edits are generally of higher quality than those of Translators. The HTER metric showed that expert NLP experts made more edits than Translators, which was further confirmed by the analysis of post-editing attributes from the human evaluation.

As for errors made and their severity scores, automatic metrics provide some clues, but this is based solely on feedback from post-editors, and not on MT and PE evaluation itself. According to post-editors, the most frequent and most serious errors in MT are terminology errors, while language errors are less frequent and associated with lower levels of severity. In this respect, human evaluation provides more precise and concrete findings. Analysis of the annotations shows that terminology errors account for a substantial proportion of both MT and PE, with a higher percentage in MT. Among post-editors, Translators make more terminology errors than NLP experts. DeepL and eTranslation have a fairly similar profile in terms of error type distribution: around 1/4 of errors are content transfer errors, and 3/4 of errors are language errors (including terminology). Systran, on the other hand, makes more language and terminology errors (around 90%) and less content transfer errors (around 10%). As a result, Systran seems to generate translations that are more correct in terms of content and meaning than DeepL and eTranslation, which deliver MT that is more linguistically and terminologically accurate. However, eTranslation is the only system that is subject to hallucinations (tool-related errors), a phenomenon that is by definition serious, which explains its poorer overall quality. Among post-editors, NLP experts produce more linguistically and terminologically correct post-edits than Translators. On the other hand, Translators produce post-edits that are more faithful to the meaning and content of the source texts than NLP experts.

In view of the considerations set out above, we can confirm that, generally speaking, the two evaluation methods reflect similar trends, but the human evaluation, which is more time-consuming and cognitively demanding, allows for a more detailed analysis and more precise results. Automatic metrics, but above all human evaluation, help compare the different translation systems and determine the profiles of the two groups of post-editors.

## 7. Conclusion

The human evaluation of this experiment consisted of a qualitative analysis of MT and PE. Using an error typology adapted for this study and a corpus of abstracts of scientific publications, as well as MT and PE compiled for this study, around twenty texts were annotated, providing valuable insight into the profiles of different MT systems and post-editors. However, for this work, only the abstracts post-edited by the two groups of post-editors were annotated, leaving out several hundred texts from the corpus that were post-edited by only one group of post-editors. The results obtained with this analysis could therefore be corroborated or nuanced by annotating a larger quantity of texts from the corpus, enabling a comparison of two evaluation methods based on exactly identical data.

Our experience shows that human evaluation remains time-consuming and cognitively demanding, and that metrics can partly compensate for this shortcoming.[7] However, our study also highlights the limitations of automatic metrics – which enable quantitative analysis – compared with human evaluation methods – which allow qualitative, explainable and interpretable analysis. An interesting approach here would be to develop a hybrid evaluation framework combining these different evaluation methods, incorporating LLM-based evaluation techniques. Large language models (LLMs) are "massive neural networks that have been trained to perform text-completion tasks on vast quantities of human-produced text" (Frankish, 2024, p. 55). This research area in NLP is starting to become an important focus in translation evaluation studies, and a number of studies have already shown encouraging results with LLMs as translation evaluation tools (cf. e.g. Fernandes et al., 2023; Kocmi & Federmann, 2023a, 2023b; Lu et al., 2023; Lu et al., 2024). However, to the best of our knowledge, to date, there has been no focus given to annotating translations based on an error typology using LLM-based methods in a practical translation teaching context.

**Acknowledgments**
This research was funded by the French Agence Nationale de la Recherche (ANR) under the project MaTOS - "ANR-22-CE23-0033-03".

---

[7] During the post-editing process by both groups, the time spent by the post-editors was recorded. However, we were unable to verify whether the post-edits were carried out as requested, i.e. without interruption and in one go. We therefore decided not to use post-editing time as a metric, as it may be biased.

**Appendix 1**

1. Content transfer
    1.1. Omission {TR-OM}
    1.2. Addition {TR-AD}
    1.3. Distortion {TR-DI}
    1.4. Indecision {TR-IN}
    1.5. Source-language-intrusion
        1.5.1. Untranslated-translatable {TR-SI-UT}
        1.5.2. Too-literal {TR-SI-TL}
        1.5.3. Units-of-weight-measurement-dates-numbers {TR-SI-UN}
    1.6. Target-language-intrusion
        1.6.1. Translated DNT {TR-TI-TD}
        1.6.2. Too-free {TI-TF}
2. Language
    2.1. Syntax {LA-SY}
        2.1.1. Determiners {LA-SY-DET}
        2.1.2. Wrong-preposition {LA-SY-PR}
        2.1.3. Complex-NP {LA-SY-GNC}
    2.2. Inflection-agreement
        2.2.1. Tense-aspect-voice {LA-IA-TA}
        2.2.2. Gender {LA-IA-GE}
        2.2.3. Number {LA-IA-NU}
    2.3. Typography
        2.3.1. Spelling {LA-HY-SP}
        2.3.2. Accents-diactritics {LA-HY-AC}
        2.3.3. Incorrect-case-upper-lower {LA-HY-CA}
        2.3.4. Punctuation {LA-HY-PU}
    2.4. Register
        2.4.1. Inconsistent-with-ST {LA-RE-IS}
        2.4.2. Inadequacy-for-TT {LA-RE-IT}
    2.5. Style
        2.5.1. Awkward {LA-ST-AW}
        2.5.2. Tautology {LA-ST-TA}
        2.5.3. Title-style {LA-ST-TS}
    2.6. Unclear-reference {LA-UR}
    2.7. Textual-conventions
        2.7.1. Coherence {LA-TC-CE}
        2.7.2. Cohesion {LA-TC-CN}
    2.8. Terminology-and-lexis
        2.8.1. Incorrect-choice-terminology {LA-TL-INS}
        2.8.2. Incorrect-choice-lexis {LA-TL-ING}
        2.8.3. Incorrect-abbreviation-acronym {LA-TL-MAA}
        2.8.4. False-cognate {LA-TL-FC}
        2.8.5. Term-translated-by-non-term {LA-TL-NT}
        2.8.6. Inappropriate-collocation-SP {LA-TL-ICS}
        2.8.7. Inappropriate-collocation-GL {LA-TL-ICG}
        2.8.8. Inconsistent-with-TT {LA-TL-IT}
        2.8.9. Terminological-inconsistency
            2.8.9.1. Different-terms-in-translation {LA-TL-TI-DT}
            2.8.9.2. Different-abbreviations-in-translation {LA-TL-TI-DA}
3. Tools.
    3.1. Hallucination {OU-TAH}
    3.2. Corpus-conformance {OU-CC}
    3.3. Duplication {OU-DU}
    3.4. Incompatible-with-glossary {OU-GC}

**Figure 7.** Error typology used in our experiment

**Appendix 2**

**Table 11.** Ranking of the 10 most frequent errors for all MT systems and post-editor groups, with (1) being the most frequent error; the errors on a grey background are the errors that are among the most frequent for the 5 variables

| Error types \ Variable | DeepL MT | Systran MT | eTranslation MT | NLP experts | Translators |
|---|---|---|---|---|---|
| Awkward style | • (6) | • (3) | • (10) | • (2) | • (7) |
| Cohesion | | • (9) | | | |
| Complex noun phrase | • (3) | • (8) | • (1) | • (8) | • (1) |
| Determination | • (5) | • (5) | | • (7) | • (4) |
| Distortion | • (4) | | • (3) | • (5) | • (3) |
| DNT translated | • (9) | | | • (6) | |
| Inappropriate register for target text | | • (2) | | | |
| Incorrect specialised collocation | • (10) | | | • (9) | |
| Incorrect terminological choice | • (1) | • (4) | • (2) | • (1) | • (2) |
| MT hallucination | | | • (7) | | |
| Number agreement | | | | | • (10) |
| Omission | | | | • (3) | |
| Punctuation | | | | • (10) | |
| Spelling | | | • (6) | | |
| Term translated by non-term | • (2) | • (1) | • (4) | | • (5) |
| Too literal | • (8) | • (6) | • (8) | • (4) | • (6) |
| Untranslated translatable | • (7) | • (7) | • (9) | | • (8) |
| Wrong preposition | | • (10) | • (5) | | • (9) |

The following definitions are those suggested in our annotation manual, available in French in Bénard et al. (2024, pp. 43-82). The annotation manual is a guide we developed in parallel with this experiment, which describes our error typology and annotation principles, and defines and illustrates the different types of error.

| | |
|---|---|
| *Awkward style* | Error characterised by unidiomatic word choice or sentence structure, giving an artificial or unnatural impression in the target language. |
| *Cohesion* | Error related to specific cohesion-related linguistic marks, such as coordination, logical connectors, anaphora, etc. (cf. e.g. Benali, 2012; Halliday and Hasan, 2014). |
| *Complex noun phrase* | Error linked to incorrect identification of the head of a noun phrase or incorrect factorisation of the various elements of a complex noun phrase. |
| *Determination* | Error related to the non-use or misuse of determiners. |
| *Distortion* | Error that alters the meaning or understanding of the source message. |
| *DNT translated* | Error related to the translation of an element into the target language when it should not be translated (DNT, Do not translate). |
| *Inappropriate register for target text* | Error linked to a register that does not comply with the register expected for the text type in question (for example, if the tone used is too informal for a scientific article). |
| *Incorrect specialised collocation* | Error in the translation of a specialised collocation (a collocation is a "combination of two or more lexical units at least one of which is a term and all the components of which do not designate one and the same concept" (Brisson, 2019, translated into English)), i.e. a collocation containing a term from the field of specialisation. |
| *Incorrect terminological choice* | Error related to the translation of a specialised term with an incorrect term in the target language. |
| *MT hallucination* | Error related to an MT output that is totally disconnected from the source text, e.g. due to incorrect factorisation or incorrect segmentation of lexical units (cf. e.g. Hansen and Esperança-Rodier, 2022). |
| *Number agreement* | Error related to number agreement. |
| *Omission* | Error related to the lack of translation of an element present in the source text. |
| *Punctuation* | Error related to punctuation (omitted or superfluous comma, missing full stop, missing space, problem with punctuation marks, etc.). |
| *Spelling* | Error concerning non-compliance with established spelling norms and rules. |
| *Term translated by non-term* | Error concerning the translation of a specialised term by an element that is not a specialised term. |
| *Too literal* | Translation error characterised by the translation being too similar to the source text, rendering the text unnatural. |
| *Untranslated translatable* | Error involving the non-translation of a term that could/should have been translated. |
| *Wrong preposition* | Error caused by the use of the wrong preposition or the non-use of a preposition. |